\documentclass{article}

\usepackage[dblblindworkshop,final]{neurips_2026}
\workshoptitle{Foundation Models for Temporal Systems}

\usepackage[utf8]{inputenc}
\usepackage[T1]{fontenc}
\usepackage[hidelinks]{hyperref}
\usepackage{url}
\usepackage{booktabs}
\usepackage{amsfonts}
\usepackage{amsmath}
\usepackage{amssymb}
\usepackage{amsthm}
\usepackage{nicefrac}
\usepackage{microtype}
\usepackage{graphicx}
\usepackage{enumitem}
\usepackage{xcolor}

\newtheorem{theorem}{Theorem}
\newtheorem{lemma}[theorem]{Lemma}
\newtheorem{proposition}[theorem]{Proposition}
\theoremstyle{remark}
\newtheorem{remark}[theorem]{Remark}

\newcommand{\Kset}{\mathcal{K}}
\newcommand{\SC}{\mathrm{SC}}
\newcommand{\Rd}{R_d}

\title{When Is Tracking Affordable? A Movement-Budget\\Feasibility Threshold for Dynamic Regret}

\author{%
  Ridham Patel \\
  Computer Science \& Engineering \\
  Indian Institute of Technology Gandhinagar \\
  Gandhinagar, India \\
  \texttt{ridham.patel@iitgn.ac.in}
}

\begin{document}

\maketitle

\begin{abstract}
Dynamic regret measures an online learner against a moving comparator, and
standard analyses assume the learner may move freely at every round.
We study what happens when it may not: the learner is given a hard budget $S$
on its total movement $\sum_t \|x_t - x_{t-1}\|$.
Our main result is that a movement budget creates a feasibility barrier.
On a domain of diameter $D$ with $G$-Lipschitz convex losses, for every
comparator path length $P_T \in [D, DT]$ we exhibit a deterministic adversary
that forces $\Rd = \Omega(GDT)$ on \emph{every}
budget-feasible algorithm whenever $S \le c P_T$ for a universal
constant $c>0$.
This barrier is caused by the budget rather than by a volatile environment:
for fixed $D$, it applies when $P_T = o(T)$, where the unconstrained minimax rate
$\Theta(G\sqrt{D(D+P_T)T})$ is sublinear, so the same instances are learnable
with an unbounded budget and unlearnable without one.
Conversely, projected online gradient descent with a budget-aware step size is
budget-feasible and attains $o(GDT)$ once $S = \omega(D + P_T)$.
Thus the critical movement scale is $D+P_T$: linear regret is unavoidable
below a sufficiently small constant multiple of $P_T$, while sublinear regret
is achievable when $S/(D+P_T)\to\infty$ in the learnable regime.
The construction rests on a distinction we make explicit: what makes a
comparator expensive to track under a movement budget is the distance it
forces the learner to travel, not its path length.
Above the threshold we give a budget-dependent upper bound and identify the
tight minimax rate in this region as an open problem rather than claiming a
product law.
Experiments support the predicted transition location and the $1/S$ decay.
\end{abstract}

\section{Introduction}

Dynamic regret in online convex optimization compares a learner's decisions
$x_t$ to a moving comparator $u_t$~\citep{zinkevich2003online,hazan2016introduction}.
Without movement constraints, the key quantity is the comparator path length
$P_T=\sum_{t\ge2}\|u_t-u_{t-1}\|$: over convex $G$-Lipschitz losses on a set
of diameter $D$, the minimax rate is
$\Theta(G\sqrt{D(D+P_T)T})$~\citep{zinkevich2003online,besbes2015non}, which
is sublinear whenever $P_T=o(DT)$.

These guarantees treat movement as free. In temporal systems, however,
adaptation is often an intervention: reconfiguring a controller, changing a
served configuration, or rebalancing a portfolio consumes a limited resource.
We impose a hard budget
$\SC=\sum_{t\ge2}\|x_t-x_{t-1}\|\le S$ and ask when sublinear dynamic regret is
feasible at all.

Our main point is that path length alone is not the right hardness measure
under a movement budget. A random-walk comparator can have large path length
while staying spatially concentrated, so a nearly static learner may still do
well. The hard case is different: the comparator must force the learner to
travel across the domain. We capture this with a deterministic full-range
square wave that alternates between the endpoints of the decision set. The
learner knows the sequence; the obstruction is purely whether it can afford to
move.

Theorem~\ref{thm:threshold} gives the resulting feasibility statement. For
every $P_T\in[D,DT]$, the square-wave instance forces
$\Rd=\Omega(GDT)$ on every learner with $S\le cP_T$, for a universal constant
$c>0$. Conversely, projected OGD with a budget-aware step size is feasible and
achieves $o(GDT)$ once $S=\omega(D+P_T)$. Thus a hard movement budget can make
tracking impossible even when, for fixed $D$ and $P_T=o(T)$, the unconstrained
problem is learnable. Appendix~\ref{app:additional} gives budget-dependent
rates above the threshold, experiments, related work, limitations, and full
proofs.

\section{Problem Setup}
\label{sec:setup}

\paragraph{Protocol.}
Fix a horizon $T$ and a compact convex decision set
$\Kset \subset \mathbb{R}^d$ with diameter
$D = \sup_{x,y \in \Kset}\|x-y\|$.
At each round $t = 1,\dots,T$ the learner selects $x_t \in \Kset$, then
observes a convex loss $f_t : \Kset \to \mathbb{R}$ that is $G$-Lipschitz,
and incurs $f_t(x_t)$.
All norms are Euclidean.

\paragraph{Comparator and path length.}
A comparator sequence is any $\{u_t\}_{t=1}^T \subset \Kset$; we take
$u_t \in \arg\min_{x \in \Kset} f_t(x)$ unless stated otherwise.
Its path length is
\begin{equation}
P_T \;=\; \sum_{t=2}^{T}\|u_t - u_{t-1}\|,
\end{equation}
which measures how far the comparator travels in total.
Note $P_T$ can be large while the comparator stays in a small region; this
range-versus-path distinction is what motivates the full-range construction
below.

\paragraph{Dynamic regret and movement budget.}
Dynamic regret and switching cost are
\begin{equation}
\Rd \;=\; \sum_{t=1}^{T}\big[f_t(x_t) - f_t(u_t)\big],
\qquad
\SC \;=\; \sum_{t=2}^{T}\|x_t - x_{t-1}\| .
\end{equation}
We call an algorithm \emph{budget-feasible} for budget $S$ if it guarantees
$\SC \le S$ on every loss sequence.
Throughout, ``linear'' regret means $\Rd = \Omega(GDT)$ and ``sublinear''
means $\Rd = o(GDT)$; the normalization by $GD$ is natural because
$|f_t(x)-f_t(y)| \le G\|x-y\| \le GD$ gives the trivial bound
$\Rd \le GDT$ for any learner.

\paragraph{Unconstrained benchmark.}
Without a budget, the minimax dynamic regret is
$\Theta(G\sqrt{D(D+P_T)T})$~\citep{zinkevich2003online,besbes2015non}, and
projected online gradient descent attains it.
This is $o(GDT)$ precisely when $P_T = o(DT)$.
For fixed $D$, the regime $P_T = o(T)$ therefore matters for us: it is where the
unconstrained problem is learnable, and hence where any impossibility result
can be attributed to the budget rather than to the environment.

\paragraph{The budgeted problem.}
Given $S$, we ask for which $S$ some budget-feasible algorithm achieves
$\Rd = o(GDT)$, and what rate is attainable when one does.

\section{The Movement-Budget Feasibility Threshold}
\label{sec:threshold}

Path length can overstate hardness under a movement budget: a random-walk
comparator has path length $\Theta(T\sigma)$ but range only
$\Theta(\sigma\sqrt T)$, so it need not force the learner to travel far.
We instead use a deterministic full-range comparator. Let
$\Kset=[0,D]$ and $f_t(x)=G|x-u_t|$. For a target $P_T\in[D,DT]$, choose
$n=\min\{\lfloor P_T/D\rfloor,T-1\}$ endpoint crossings and let $u_t$
alternate between $0$ and $D$ in $n+1$ long blocks. The realized path length
is at most $P_T$, and the learner knows the comparator exactly; the only
obstacle is whether it can afford the crossings.

\begin{theorem}[Movement-budget feasibility threshold]
\label{thm:threshold}
Fix a horizon $T$, a domain of diameter $D$, and a path length
$P_T$ with $D \le P_T \le DT$.
\begin{enumerate}[leftmargin=1.6em,itemsep=1pt,topsep=2pt]
\item \textbf{(Impossibility.)}
There exist universal constants $c_0,c_1>0$ such that for every algorithm
with $\SC \le S$ there is a comparator sequence of path length at most
$P_T$ and a sequence of convex, $G$-Lipschitz losses on which
\begin{equation}
\Rd \;\ge\; c_0 GDT\left(1 - c_1\frac{S}{P_T}\right).
\label{eq:lb}
\end{equation}
In particular $\Rd = \Omega(GDT)$ whenever $S \le cP_T$ for a sufficiently
small universal constant $c>0$.
\item \textbf{(Achievability.)}
If $S = \omega(D+P_T)$ and $S = o(DT)$, then projected OGD with
$\eta = S/(GT)$ satisfies $\SC \le S$ and $\Rd = o(GDT)$.
\end{enumerate}
Thus the critical movement scale is $D+P_T$: below a sufficiently small
constant multiple of $P_T$ linear regret is unavoidable, while
$S/(D+P_T)\to\infty$ is sufficient for sublinear regret in the learnable
regime. Since the theorem assumes $P_T\ge D$, this identifies $P_T$ as the
asymptotic movement scale, without resolving the exact constant-factor band.
\end{theorem}

The lower bound is a counting argument over consecutive square-wave block
pairs: on each pair, if the learner spends movement $M_i$, its cumulative
tracking error is at least a constant multiple of the block length times
$(D-M_i)$. Summing over $\Theta(P_T/D)$ pairs and using
$\sum_i M_i\le S$ gives \eqref{eq:lb}; Appendix~\ref{app:thm1} gives the
full proof with floor and ceiling effects. The constructive direction uses
the standard OGD bound
$\Rd\le D(D+2P_T)/(2\eta)+\eta G^2T/2$~\citep{zinkevich2003online}. With
$\eta=S/(GT)$, movement is at most $T\eta G=S$ and
\begin{equation}
\Rd \le \frac{GTD(D+2P_T)}{2S}+\frac{GS}{2},
\label{eq:ub-constrained}
\end{equation}
which is $o(GDT)$ when $S=\omega(D+P_T)$ and $S=o(DT)$.

\begin{figure}[t]
\centering
\includegraphics[width=0.78\textwidth]{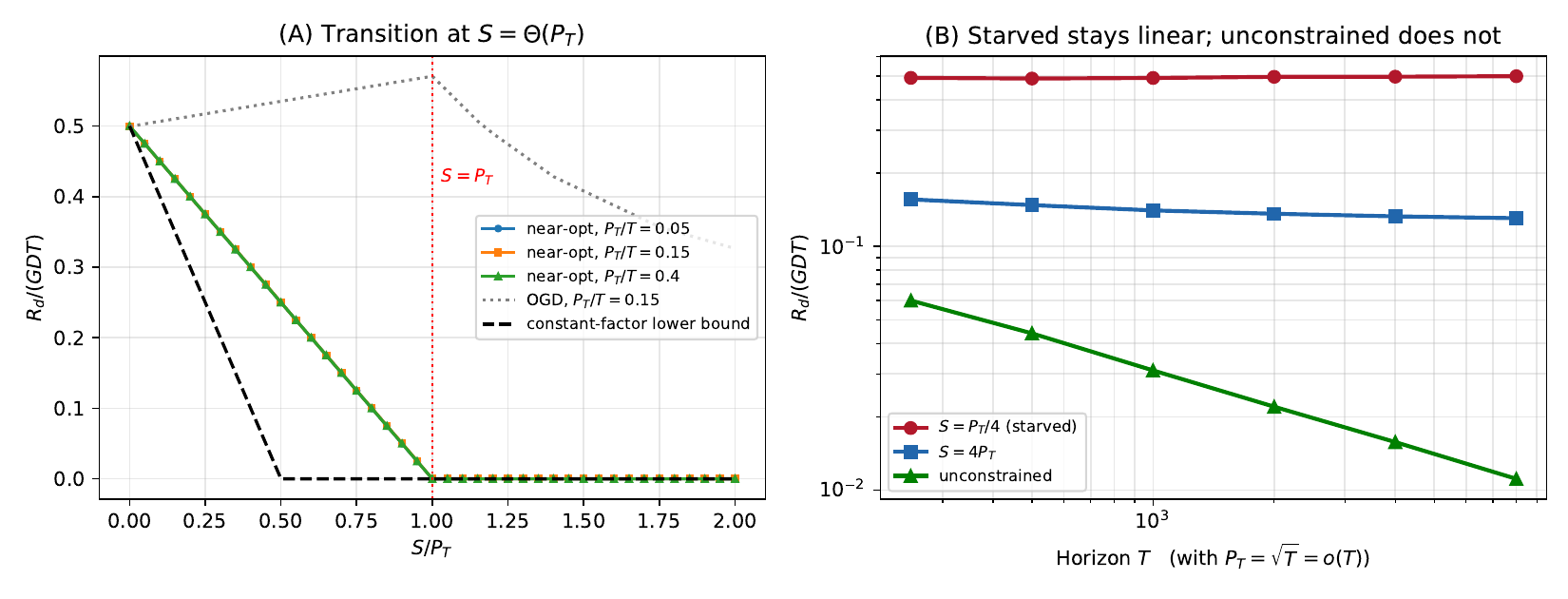}
\caption{Square-wave adversary. Left: normalized regret drops as the movement
budget approaches the comparator path length; the proved lower bound has only
constant-factor precision. Right: with $P_T=\sqrt T=o(T)$, a budget-starved
learner has constant normalized regret while the unconstrained learner decays,
showing that the barrier is caused by the movement budget rather than by
large path length.}
\label{fig:squarewave-main}
\end{figure}

For fixed $D$, the theorem is nonvacuous at $P_T=o(T)$: the unconstrained
rate $\Theta(G\sqrt{D(D+P_T)T})$ is then $o(GDT)$, yet any learner with
$S\le cP_T$ suffers linear regret on the same type of instance. Figure
\ref{fig:squarewave-main} illustrates this transition and the sublinear-path
regime; additional experiments and the remaining rate gap above the threshold
are in Appendix~\ref{app:additional}.


\section{Rates Above the Threshold and Evidence}
\label{sec:main-rates-evidence}

The feasibility theorem answers when sublinear regret is possible. Above that
threshold, the finer rate is only partially characterized. Running projected
OGD with
\[
\eta=\min\!\left\{\sqrt{\frac{D(D+2P_T)}{G^2T}},\frac{S}{GT}\right\}
\]
keeps $\SC\le S$ and gives
\begin{equation}
\Rd\le O\!\left(\min\left\{G\sqrt{D(D+2P_T)T}+\frac{GTD(D+2P_T)}{S},\,GDT\right\}\right).
\label{eq:main-rate}
\end{equation}
The first term is the usual unconstrained OGD rate; the second is the price
of shrinking the step size to fit the movement budget. The standard
unconstrained lower bound
$\Omega(G\sqrt{DP_TT})$~\citep{besbes2015non} still applies, but we do not
claim a matching lower bound for the budget-dependent $1/S$ term. Determining
the minimax rate above the feasibility threshold remains open.

Our experiments are designed around this separation. Figure~\ref{fig:squarewave-main}
tests the theorem itself: the left panel shows the transition as $S/P_T$ varies,
while the right panel fixes $P_T=\sqrt T=o(T)$ and separates a budget-induced
barrier from ordinary nonstationarity. Appendix~\ref{sec:experiments} reports
two further checks. First, random-walk and rotating-comparator experiments show
an empirical $1/S$ decay consistent with the upper bound \eqref{eq:main-rate},
but we do not treat this as proof that the $1/S$ term is minimax tight. Second, five real
daily series, S\&P~500, EUR/USD, crude oil via USO, Bitcoin, and New York City
temperature, show regret remaining high for $S\ll P_T$ and dropping near
$S=P_T$. These finite-horizon experiments test transition location and
budget--regret behavior; asymptotic linearity is certified by the square-wave
construction, not by the real series.

The closest prior work studies adjacent but different questions. Classical
dynamic-regret results characterize adaptation to a moving comparator when
movement is free~\citep{zinkevich2003online,besbes2015non}, with later work
removing prior knowledge of $P_T$ or improving constants in richer loss
classes~\citep{yang2017tracking,baby2021optimal,zhang2022sword}. Smoothed
online optimization penalizes movement with a Lagrange multiplier rather than
imposing a hard budget~\citep{lin2011dynamic,andrew2013tale,chen2018smoothed,goel2019beyond,zhang2021revisiting}. Hard
switching-budget work is closest in spirit, but existing results focus on
static regret or switching count rather than a distance budget for dynamic
regret~\citep{dekel2014bandits,chen2020minimax,sherman2021lazy,zhang2021continuous}. Our contribution is the
feasibility barrier: even when unconstrained dynamic regret is sublinear, a
hard movement budget below the scale $D+P_T$ can make tracking impossible.

This perspective is useful for temporal systems because it separates two
roles often conflated under the term ``adaptation.'' Prediction asks whether
the learner can infer where the moving optimum is; intervention asks whether
the learner can afford to move there. The square-wave construction removes
the prediction issue entirely: the comparator is deterministic and known, yet
the learner still fails when its movement budget is too small. Thus the
threshold is not a statistical limitation but a physical or operational one.

The remaining gap is also conceptually sharp. Below the threshold, range
forces a feasibility barrier. Far above it, standard OGD rates reappear. In
between, the known upper bound has a $1/S$ term, but no matching
budget-dependent lower bound is known. Deciding whether this term is an OGD
artifact or an unavoidable cost is the main technical question left by the
paper.

\bibliographystyle{plainnat}
\bibliography{references}

@inproceedings{zinkevich2003online,
  title={Online convex programming and generalized infinitesimal gradient ascent},
  author={Zinkevich, Martin},
  booktitle={Proceedings of the 20th international conference on machine learning (icml-03)},
  pages={928--936},
  year={2003}
}

@article{hazan2016introduction,
  title={Introduction to online convex optimization},
  author={Hazan, Elad},
  journal={Foundations and Trends in Optimization},
  volume={2},
  number={3-4},
  pages={157--325},
  year={2016},
  publisher={Emerald Publishing Limited}
}

@inproceedings{chen2018smoothed,
  title={Smoothed online convex optimization in high dimensions via online balanced descent},
  author={Chen, Niangjun and Goel, Gautam and Wierman, Adam},
  booktitle={Conference On Learning Theory},
  pages={1574--1594},
  year={2018},
  organization={PMLR}
}

@INPROCEEDINGS{lin2011dynamic,
  author={Lin, Minghong and Wierman, Adam and Andrew, Lachlan L. H. and Thereska, Eno},
  booktitle={2011 Proceedings IEEE INFOCOM}, 
  title={Dynamic right-sizing for power-proportional data centers}, 
  year={2011},
  volume={},
  number={},
  pages={1098-1106},
  doi={10.1109/INFCOM.2011.5934885}}

@inproceedings{andrew2013tale,
  title={A tale of two metrics: Simultaneous bounds on competitiveness and regret},
  author={Andrew, Lachlan and Barman, Siddharth and Ligett, Katrina and Lin, Minghong and Meyerson, Adam and Roytman, Alan and Wierman, Adam},
  booktitle={Conference on Learning Theory},
  pages={741--763},
  year={2013},
  organization={PMLR}
}

@article{zhang2021continuous,
  title={Online convex optimization with continuous switching constraint},
  author={Wang, Guanghui and Wan, Yuanyu and Yang, Tianbao and Zhang, Lijun},
  journal={Advances in Neural Information Processing Systems},
  volume={34},
  pages={28636--28647},
  year={2021}
}

@article{besbes2015non,
  title={Non-stationary stochastic optimization},
  author={Besbes, Omar and Gur, Yonatan and Zeevi, Assaf},
  journal={Operations research},
  volume={63},
  number={5},
  pages={1227--1244},
  year={2015},
  publisher={INFORMS}
}

@article{zhang2018adaptive,
  title={Adaptive online learning in dynamic environments},
  author={Zhang, Lijun and Lu, Shiyin and Zhou, Zhi-Hua},
  journal={Advances in neural information processing systems},
  volume={31},
  year={2018}
}

@article{zhao2020dynamic,
  title={Dynamic regret of convex and smooth functions},
  author={Zhao, Peng and Zhang, Yu-Jie and Zhang, Lijun and Zhou, Zhi-Hua},
  journal={Advances in Neural Information Processing Systems},
  volume={33},
  pages={12510--12520},
  year={2020}
}

@article{zhang2021revisiting,
  title={Revisiting smoothed online learning},
  author={Zhang, Lijun and Jiang, Wei and Lu, Shiyin and Yang, Tianbao},
  journal={Advances in Neural Information Processing Systems},
  volume={34},
  pages={13599--13612},
  year={2021}
}

@article{chen2020minimax,
  title={Minimax regret of switching-constrained online convex optimization: No phase transition},
  author={Chen, Lin and Yu, Qian and Lawrence, Hannah and Karbasi, Amin},
  journal={Advances in Neural Information Processing Systems},
  volume={33},
  pages={3477--3486},
  year={2020}
}

@inproceedings{dekel2014bandits,
  title={Bandits with switching costs: {$T^{2/3}$} regret},
  author={Dekel, Ofer and Ding, Jian and Koren, Tomer and Peres, Yuval},
  booktitle={Proceedings of the forty-sixth annual ACM symposium on Theory of computing},
  pages={459--467},
  year={2014}
}

@article{jacobsen2024equivalence,
  title={An equivalence between static and dynamic regret minimization},
  author={Jacobsen, Andrew and Orabona, Francesco},
  journal={Advances in Neural Information Processing Systems},
  volume={37},
  pages={98835--98870},
  year={2024}
}

@article{mezic2005spectral,
  title={Spectral properties of dynamical systems, model reduction and decompositions},
  author={Mezi{\'c}, Igor},
  journal={Nonlinear Dynamics},
  volume={41},
  number={1},
  pages={309--325},
  year={2005},
  publisher={Springer}
}

@article{budisic2012applied,
  title={Applied koopmanism},
  author={Budi{\v{s}}i{\'c}, Marko and Mohr, Ryan and Mezi{\'c}, Igor},
  journal={Chaos: An Interdisciplinary Journal of Nonlinear Science},
  volume={22},
  number={4},
  year={2012},
  publisher={AIP Publishing}
}

@book{daley2007introduction,
  title={An introduction to the theory of point processes: volume II: general theory and structure},
  author={Daley, Daryl J and Vere-Jones, David},
  year={2008},
  publisher={Springer}
}

@inproceedings{baby2021optimal,
  title={Optimal dynamic regret in exp-concave online learning},
  author={Baby, Dheeraj and Wang, Yu-Xiang},
  booktitle={Conference on Learning Theory},
  pages={359--409},
  year={2021},
  organization={PMLR}
}

@article{zhang2022sword,
    author = {Peng Zhao and
                    Yu-Jie Zhang and
                    Lijun Zhang and
                    Zhi-Hua Zhou},
    title = {Adaptivity and Non-stationarity: Problem-dependent Dynamic Regret for Online Convex Optimization},
    journal = {Journal of Machine Learning Research},
    volume = {25},
    number = {98},
    pages = {1--52},
    year = {2024}
}

@inproceedings{sherman2021lazy,
  title={Lazy oco: Online convex optimization on a switching budget},
  author={Sherman, Uri and Koren, Tomer},
  booktitle={Conference on Learning Theory},
  pages={3972--3988},
  year={2021},
  organization={PMLR}
}

@inproceedings{yang2017tracking,
  title={Tracking slowly moving clairvoyant: Optimal dynamic regret of online learning with true and noisy gradient},
  author={Yang, Tianbao and Zhang, Lijun and Jin, Rong and Yi, Jinfeng},
  booktitle={International Conference on Machine Learning},
  pages={449--457},
  year={2016},
  organization={PMLR}
}

@article{goel2019beyond,
  title={Beyond online balanced descent: An optimal algorithm for smoothed online optimization},
  author={Goel, Gautam and Lin, Yiheng and Sun, Haoyuan and Wierman, Adam},
  journal={Advances in Neural Information Processing Systems},
  volume={32},
  year={2019}
}

\newpage
\appendix

\section{Additional Rates, Experiments, and Discussion}
\label{app:additional}

We state this as a threshold in \emph{scale} and not as a sharp threshold
with a determined constant, and the distinction is real.
The impossibility bound \eqref{eq:lb} has only universal-constant precision,
while the experiments in Section~\ref{sec:exp-sqwave} indicate that the
empirical transition is close to $S=P_T$.
Section~\ref{sec:exp-sqwave} reports the empirical constant separately from
the proved one, and we do not conflate them.

\section{Dynamic Regret Above the Threshold}
\label{sec:rates}

Above the threshold the question becomes quantitative.
The results in this section refine Theorem~\ref{thm:threshold} and are
secondary to it; in particular they do not resolve the rate in the whole
range of budgets.

\begin{theorem}[Budget-dependent upper bound]
\label{thm:rates}
Assume each $f_t$ is convex and $G$-Lipschitz on $\Kset$.
Projected OGD with
$\eta = \min\!\big(\sqrt{D(D+2P_T)/(G^2T)},\; S/(GT)\big)$
satisfies $\SC \le S$ and
\begin{equation}
\Rd \;\le\;
O\!\left(\min\left\{
G\sqrt{D(D+2P_T)T} + \frac{GTD(D+2P_T)}{S},\;\; GDT
\right\}\right).
\label{eq:rate-ub}
\end{equation}
\end{theorem}

The usual unconstrained lower bound
$\Rd \ge \Omega(G\sqrt{DP_TT})$~\citep{besbes2015non} applies to this
setting as well, since adding a movement budget cannot make the problem
easier. We do not know whether the additional $1/S$ term in
\eqref{eq:rate-ub} is minimax tight.

\medskip
\begin{remark}[What is and is not established]
\label{rem:gap}
Theorem~\ref{thm:rates} is an upper bound, not a minimax characterization.
We therefore do \emph{not} claim a product law
$\Rd \cdot S = \Theta(GTD(D+P_T))$.
The exact rate above the feasibility threshold, and in particular the
tightness of the $GTD(D+2P_T)/S$ term, is open.
\end{remark}

We also note that \eqref{eq:rate-ub} is informative only when
$S \gtrsim D + P_T$: for smaller $S$ the term $GTD(D+2P_T)/S$ exceeds the
trivial bound $GDT$, and Theorem~\ref{thm:threshold} shows linear regret is
in fact unavoidable there.
The two results thus cover complementary ranges of $S$.

\section{Budget-Controlled OGD}
\label{sec:algorithm}

The algorithm used for both constructive results is projected OGD with a
budget-aware step size:
\begin{equation}
x_{t+1} \;=\; \Pi_{\Kset}\!\left(x_t - \eta\,\nabla f_t(x_t)\right),
\qquad
\eta \;=\; \min\!\left(\sqrt{\frac{D(D+2P_T)}{G^2T}},\;\; \frac{S}{GT}\right).
\label{eq:alg}
\end{equation}

\begin{proposition}[Budget-Controlled OGD]
\label{prop:ogd}
Assume each $f_t$ is convex and $G$-Lipschitz.
The update \eqref{eq:alg} satisfies $\SC \le S$ and
$\mathbb{E}[\Rd] \le O\!\big(G\sqrt{D(D+2P_T)T} + GTD(D+2P_T)/S\big)$.
When $P_T$ is unknown, a doubling trick preserves the rate up to constants.
\end{proposition}

The algorithm is not new: it is the standard OGD of
\citet{zinkevich2003online} run at a particular step size.
What the analysis contributes is the identification of that step size.
The two arguments of the minimum in \eqref{eq:alg} correspond to the two
regimes: the first is the usual unconstrained optimum, the second is the
largest step for which the budget is respected without any explicit
projection onto the constraint, since per-round movement is bounded by
$\eta G$ and there are $T$ rounds.
Budget feasibility is thus enforced by the step size alone.
We do not claim \eqref{eq:alg} is a new optimization method, and by
Remark~\ref{rem:gap} we use it as a constructive upper bound rather than as
a fully minimax-optimal method above the threshold.

\section{Experiments}
\label{sec:experiments}

The experiments are organized by which claim they test.
Section~\ref{sec:exp-sqwave} uses the square wave, the instance our lower
bound is actually proved against.
Section~\ref{sec:exp-sublinear} isolates the regime $P_T = o(T)$ that makes
the barrier attributable to the budget.
The remaining subsections use a random-walk comparator and real series to
examine the transition location and the $1/S$ decay. A random-walk comparator
is spatially concentrated, so it cannot certify a budget-induced barrier, and
we do not use it for that purpose.

\begin{figure}[t]
\centering
\includegraphics[width=\textwidth]{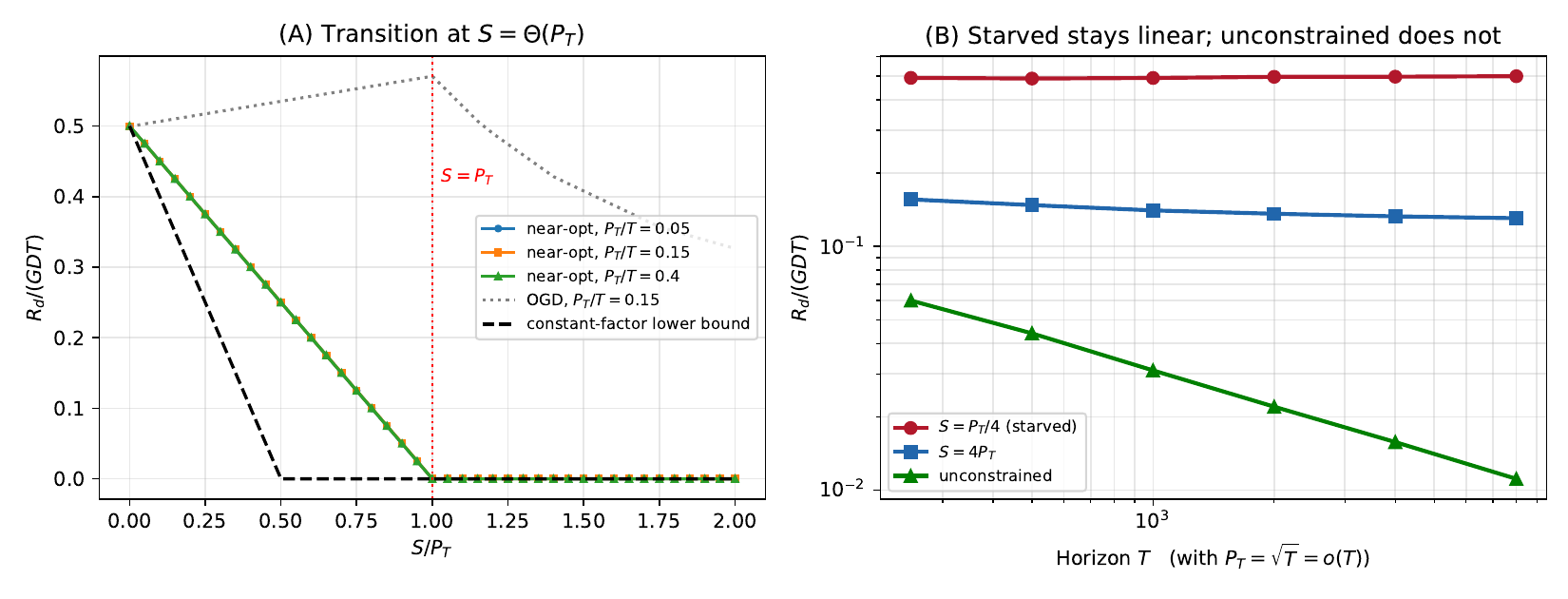}
\caption{Square-wave adversary on $\Kset=[0,D]$ with $f_t(x)=G|x-u_t|$.
\textbf{(A)} Normalized regret of a near-optimal budget-feasible tracker
against $S/P_T$.
Curves for $P_T/T \in \{0.05,0.15,0.4\}$ collapse onto
$\tfrac12(1-S/P_T)$ and reach zero at $S = P_T$.
The dashed line shows the scale of the proved lower bound in
Theorem~\ref{thm:threshold}; its constants are conservative, and the
empirical transition occurs closer to $S=P_T$.
The dotted curve is Budget-Controlled OGD, which pays an additional lag
penalty on this instance.
\textbf{(B)} Fixing $P_T = \sqrt{T} = o(T)$ and growing $T$: the starved
learner ($S = P_T/4$) stays flat near $0.5$, so its regret is linear, while
the unconstrained learner decays as $T^{-1/2}$.
Both axes logarithmic.}
\label{fig:squarewave}
\end{figure}

\subsection{The budget-induced transition}
\label{sec:exp-sqwave}

Figure~\ref{fig:squarewave}(A) sweeps $S/P_T$ for a near-optimal
budget-feasible tracker at $T = 4000$.
The curves for three path lengths collapse onto a single line
$\Rd/(GDT) = \tfrac12(1 - S/P_T)$ that vanishes at $S = P_T$.
The proved lower bound has more conservative constants and lies below this
empirical curve.

We report these two constants separately on purpose.
An exact dynamic program over discretized position and budget, which lower
bounds every algorithm, places the true optimum slightly below
$\tfrac12(1-S/P_T)$: at $T=240$ and $P_T=29$ it gives $0.364$ versus $0.375$
at $S = P_T/4$, and $0.242$ versus $0.250$ at $S = P_T/2$, in units of
$GDT$.
Crucially it vanishes only at $S = P_T$.
This is empirical evidence that the constants in
Theorem~\ref{thm:threshold} are conservative and that some slack sits in
inequality \eqref{eq:pair}.
It is not a theorem, and we do not upgrade it to one; the proved statement
remains \eqref{eq:lb}.

The dotted curve shows Budget-Controlled OGD rising above the frozen-learner
value before falling.
This is a lag effect specific to this instance: at small step sizes the
iterate arrives at one endpoint around the time the comparator leaves it, so
part of the horizon is spent in antiphase.
It does not contradict Theorem~\ref{thm:rates}, whose upper bound is a
worst-case statement and is not tight here, but it does show that the
near-optimal tracker and OGD differ by more than a constant on this family.

\subsection{The sublinear-path-length regime}
\label{sec:exp-sublinear}

Figure~\ref{fig:squarewave}(B) fixes $P_T = \sqrt T$, so $P_T = o(T)$, and
grows $T$ from $250$ to $8000$.
The budget-starved learner at $S = P_T/4$ has normalized regret flat in $T$,
moving from $0.492$ to $0.500$, so its regret is linear.
The unconstrained learner decays from $0.060$ to $0.011$ over the same range,
consistent with $T^{-1/2}$.

This is the experiment that separates the two hypotheses.
Both learners face identical loss sequences drawn from a regime where the
unconstrained problem is learnable.
One is starved of movement and is not.
This is the empirical counterpart of the nonvacuity statement in
Theorem~\ref{thm:threshold}: the same regime is learnable without a budget but
not with insufficient movement. It is the claim a random-walk comparator cannot
support.

\subsection{Budget--regret trade-off}
\label{sec:exp-tradeoff}

\begin{figure}[t]
\centering
\includegraphics[width=0.56\textwidth]{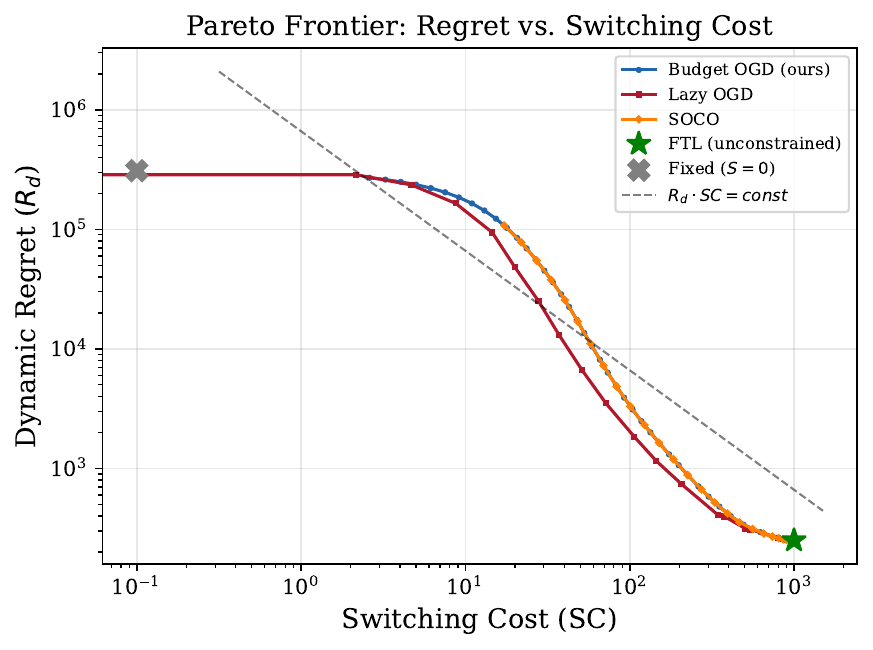}
\caption{Empirical budget--regret trade-off on a random-walk comparator.
All methods trace a $1/S$ decay over the mid-range of budgets, with
Budget-Controlled OGD attaining the lowest regret at each budget.
This is consistency with the upper bound \eqref{eq:rate-ub} only.
The dashed guide line is a $1/S$ reference and is not evidence of a matched
product law; by Remark~\ref{rem:gap} tightness of the $1/S$ term is
unproved.}
\label{fig:pareto}
\end{figure}

Figure~\ref{fig:pareto} plots the empirical $(\SC, \Rd)$ trade-off for
Budget-Controlled OGD, Lazy OGD, and SOCO on a random-walk comparator.
All three trace a $1/S$ decay over the mid-range of budgets, with
Budget-Controlled OGD attaining the lowest regret at each budget level.

This is empirical consistency with the \emph{upper} bound
\eqref{eq:rate-ub} only.
By Remark~\ref{rem:gap}, the tightness of the $1/S$ term is open, so these
curves must not be read as confirming a product law.

\begin{figure}[t]
\centering
\includegraphics[width=\textwidth]{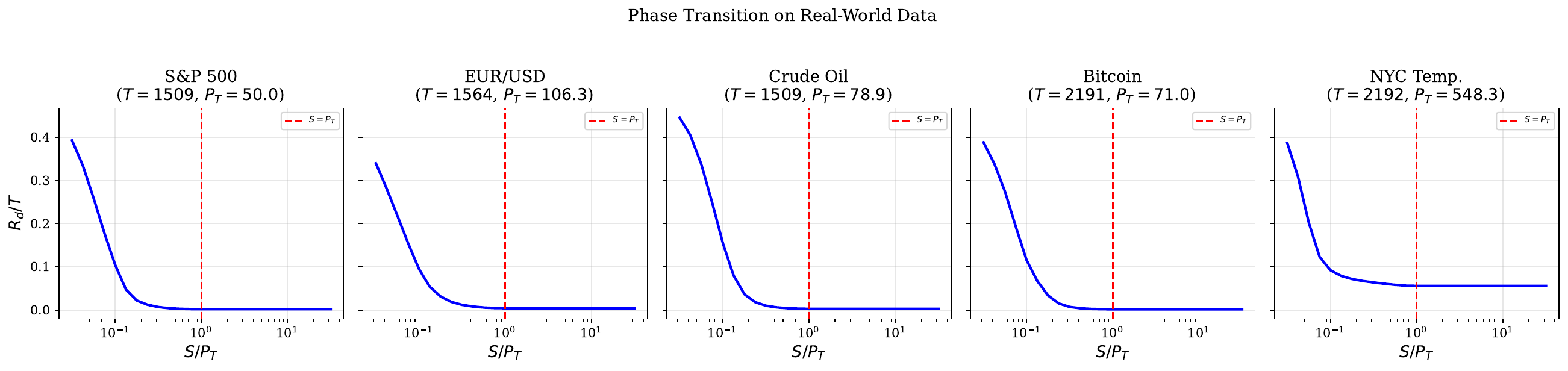}
\caption{Transition location on five real series (S\&P~500, EUR/USD, crude
oil, Bitcoin, NYC temperature).
Regret stays elevated for $S \ll P_T$ and drops near $S = P_T$ on all five,
placing the transition where Theorem~\ref{thm:threshold} predicts.
Finite horizons cannot separate $\Theta(T)$ from $o(T)$, so these panels
test the \emph{location} of the transition and the $1/S$ decay, not
asymptotic linearity; that claim rests on Figure~\ref{fig:squarewave}(B).}
\label{fig:realphase}
\end{figure}

\subsection{Algorithm comparison}
\label{sec:exp-algo}

\begin{figure}[h]
\centering
\begin{minipage}[t]{0.48\textwidth}
\centering
\includegraphics[width=\textwidth]{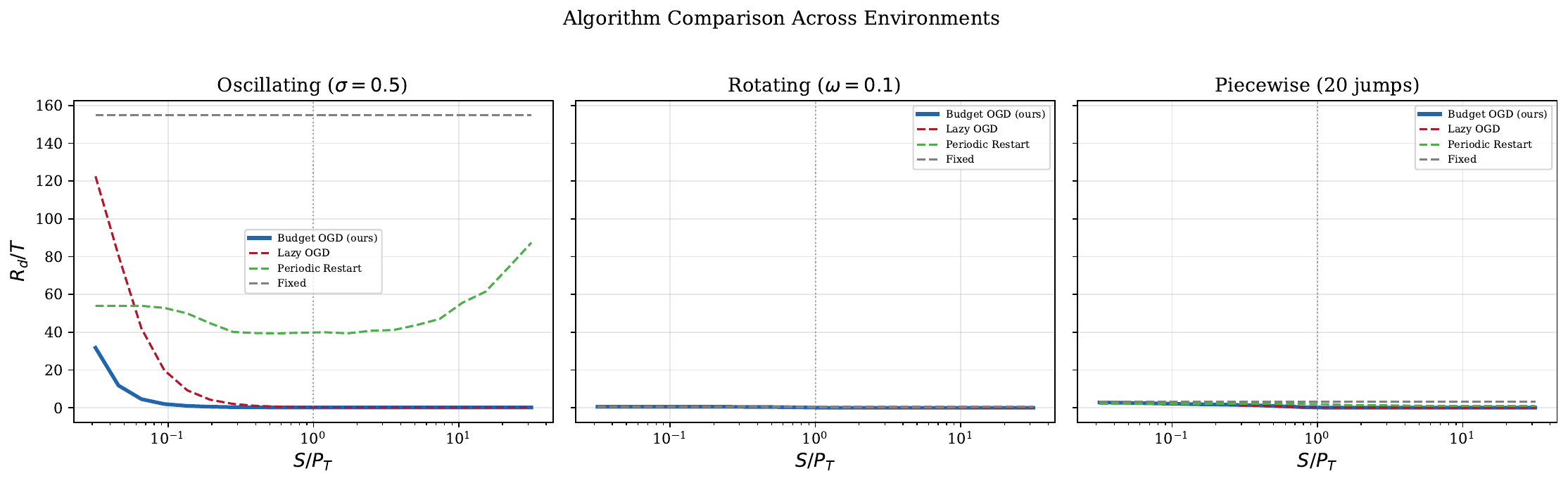}
\caption{Budget-Controlled OGD against Lazy OGD, SOCO, and Periodic Restart
across oscillating, rotating, and piecewise stationary environments. Periodic
Restart follows the standard restarting baseline for non-stationary online
optimization~\citep{besbes2015non}.
Budget-Controlled OGD matches or outperforms the baselines at all budget
levels.}
\label{fig:algo}
\end{minipage}\hfill
\begin{minipage}[t]{0.48\textwidth}
\centering
\includegraphics[width=\textwidth]{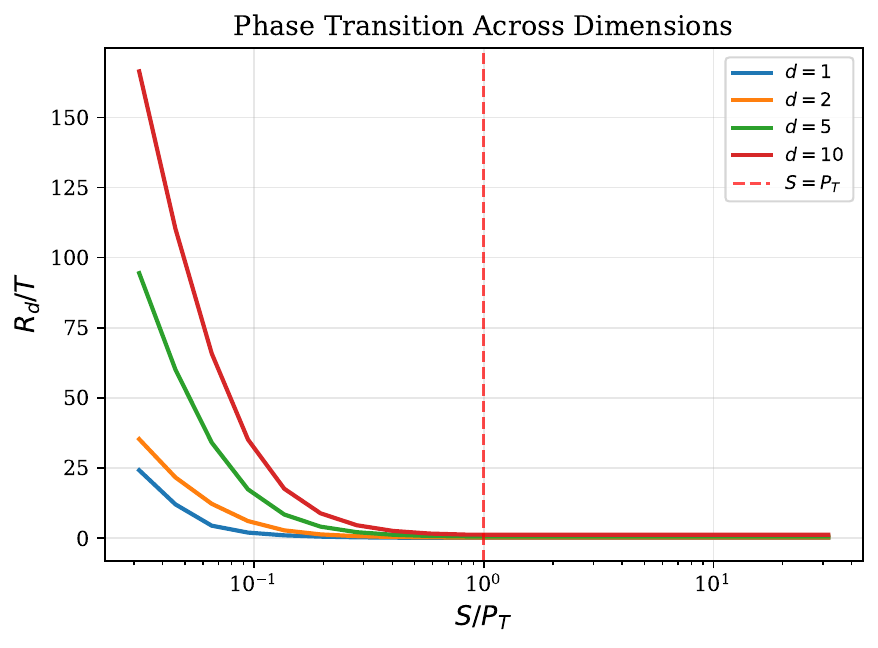}
\caption{Transition location for $d \in \{1,2,5,10,20\}$.
The location $S/P_T \approx 1$ is flat in $d$ because $S$ and $P_T$ are
measured in the same norm; how the constants and $D$ scale with dimension is
not resolved here.}
\label{fig:dim}
\end{minipage}
\end{figure}

Across oscillating, rotating, and piecewise stationary synthetic
environments, Budget-Controlled OGD matches or outperforms Lazy OGD, SOCO,
and Periodic Restart at all budget levels
(Appendix~\ref{app:extra-exp}, Figure~\ref{fig:algo}); Periodic Restart is
included as a standard baseline for non-stationary online
optimization~\citep{besbes2015non}.
Lazy OGD is handicapped by its threshold update rule, which spends budget on
small moves during low-volatility stretches; Periodic Restart suffers when
its restart boundaries are misaligned with the environment.
On real data the same ordering holds on the S\&P~500, Bitcoin, and NYC
temperature series.

\subsection{Real-world series}
\label{sec:exp-real}

We use five daily series as comparator sequences, each normalized:
the S\&P~500 index ($T{=}1509$), the EUR/USD exchange rate ($T{=}1564$),
crude oil via the USO ETF ($T{=}1509$), Bitcoin ($T{=}2191$), and mean
temperature in New York City ($T{=}2192$).

Figure~\ref{fig:realphase} shows regret staying elevated for $S \ll P_T$ and
dropping near $S = P_T$ on all five, locating the transition where
Theorem~\ref{thm:threshold} places it.
The drop is sharpest for Bitcoin and temperature and more gradual for the
S\&P~500, whose trajectory is smoother.
Trajectory-level behaviour matches: at $S = 0.05 P_T$ the learner barely
moves and incurs $\Rd/T \approx 0.47$; at $S = 0.3 P_T$ tracking improves
substantially; at $S = P_T$ the learner tracks near-perfectly with
$\Rd/T < 0.01$ (Appendix~\ref{app:extra-exp}).

\paragraph{Finite-horizon caveat.}
These series have modest path-length ratios: $P_T/T$ is $0.033$
(S\&P~500), $0.032$ (Bitcoin), $0.068$ (EUR/USD), $0.052$ (oil) and $0.250$
(temperature), against $\sqrt{T(1+P_T)}$ of $277$, $397$, $410$, $347$ and
$1097$ respectively.
Every dataset therefore sits in the regime $P_T \ll \sqrt{T(1+P_T)}$, where
the window $P_T \lesssim S \lesssim \sqrt{D(D+P_T)T}$ spans roughly half a
decade in $S$.
At a single finite horizon one cannot separate $\Theta(T)$ from $o(T)$, so
these experiments test the location of the transition and the $1/S$ decay
and cannot by themselves establish asymptotic linearity.
That claim rests on Figure~\ref{fig:squarewave}(B).

\section{Related Work}
\label{sec:related}

\paragraph{Dynamic regret.}
Dynamic regret against a moving comparator originates with
\citet{zinkevich2003online}, who gave the $O(D(D+P_T)/\eta + \eta G^2T)$
bound for OGD that we use.
\citet{besbes2015non} established minimax rates for non-stationary
stochastic optimization and supply the $\Omega(G\sqrt{DP_TT})$ lower bound
in Theorem~\ref{thm:rates}.
Subsequent work develops adaptive and parameter-free methods that attain
path-length-dependent rates without knowing
$P_T$~\citep{yang2017tracking,zhang2018adaptive,zhao2020dynamic,baby2021optimal,zhang2022sword}, and
\citet{jacobsen2024equivalence} relates static and dynamic regret through
lifted constructions.
All of this assumes movement is free.

\paragraph{Switching costs and smoothed online optimization.}
Smoothed online convex optimization augments the loss with a movement
penalty, $\sum_t f_t(x_t) + \lambda\sum_t\|x_t-x_{t-1}\|$, and yields
competitive algorithms such as online balanced
descent~\citep{lin2011dynamic,andrew2013tale,chen2018smoothed,goel2019beyond}.
Recent work also studies dynamic regret when the switching cost is part of
the performance metric~\citep{zhang2021revisiting}.
A soft penalty selects a single operating point through $\lambda$; it does
not answer whether learning is possible at a prescribed budget, which is the
question Theorem~\ref{thm:threshold} addresses.

\paragraph{Hard switching constraints.}
Closest to our setting are hard constraints on movement.
\citet{zhang2021continuous} study continuous switching distance constraints
and \citet{sherman2021lazy} constrain the switching \emph{count}; both
target static regret. Switching-budget lower bounds also appear in bandit
and switching-constrained OCO models~\citep{dekel2014bandits,chen2020minimax}.
Table~\ref{tab:comparison} summarizes the positioning.
To our knowledge no prior work identifies a feasibility threshold for
\emph{dynamic} regret under a hard movement-distance budget.

\begin{table}[t]
\centering
\small
\caption{Positioning. ``Dyn.''\ and ``Stat.''\ denote dynamic and static
regret; ``Dist.''\ and ``Count'' denote switching-distance and
switching-count constraints.}
\label{tab:comparison}
\begin{tabular}{lcccc}
\toprule
\textbf{Work} & \textbf{Regret} & \textbf{Constraint} & \textbf{Minimax} & \textbf{Threshold} \\
\midrule
\citet{besbes2015non}      & Dyn.  & None  & Yes     & No \\
\citet{baby2021optimal}    & Dyn.  & None  & Yes     & No \\
\citet{zhang2022sword}     & Dyn.  & None  & No      & No \\
\citet{zhang2021revisiting}& Dyn.  & Soft  & Yes     & No \\
\citet{dekel2014bandits}   & Stat. & Switch cost & Yes & No \\
\citet{chen2020minimax}    & Stat. & Count & Yes     & No \\
\citet{sherman2021lazy}    & Stat. & Count & Yes     & No \\
\citet{zhang2021continuous}& Stat. & Dist. & Partial & No \\
\citet{chen2018smoothed}   & Dyn.  & Soft  & No      & No \\
\midrule
\textbf{This work}         & \textbf{Dyn.} & \textbf{Dist.} & \textbf{Open}$^\dagger$ & \textbf{Yes} \\
\bottomrule
\end{tabular}

\vspace{2pt}
{\footnotesize $^\dagger$Theorem~\ref{thm:threshold} identifies $D+P_T$ as
the critical movement scale: a small constant multiple of $P_T$ is
insufficient, while $S/(D+P_T)\to\infty$ is sufficient in the learnable
regime.
The exact minimax rate above that threshold remains open
(Remark~\ref{rem:gap}).}
\end{table}

\section{Limitations and Open Problems}
\label{sec:limitations}

\paragraph{Rates above the threshold.}
The main open problem is the minimax rate once $S$ exceeds the feasibility
threshold.
Resolving it appears to require understanding whether the $1/S$ term in
Theorem~\ref{thm:rates} is an OGD artifact or reflects an unavoidable
budget-dependent price.
The two natural lower-bound constructions each fail on one count.
Our square wave is deterministic and becomes exactly trackable once
$S \ge P_T$, so it certifies nothing above the threshold.
A drifting comparator can be followed with $O(P_T)$ additional movement once
the observed losses reveal its position, so for $S \gg P_T$ increasing $P_T$
does not obviously help the adversary.
We therefore do not claim minimax optimality for Theorem~\ref{thm:rates}.

\paragraph{The constant in the threshold.}
Theorem~\ref{thm:threshold} proves the threshold only up to universal
constants, while the exact dynamic program of Section~\ref{sec:exp-sqwave}
indicates the empirical transition is close to $S=P_T$.
The result should therefore be read as identifying the asymptotic movement
scale $D+P_T$, not as determining the exact constant-factor transition.
The slack is localized to inequality \eqref{eq:pair}.

\paragraph{What each piece of evidence supports.}
We distinguish three levels deliberately.
The square-wave construction \emph{proves} linear regret for $S \le cP_T$
for a universal constant $c>0$, uniformly over $P_T \in [D,DT]$, and
Figure~\ref{fig:squarewave}(B) provides
the corresponding empirical demonstration at $P_T = o(T)$.
The random-walk and real-data experiments \emph{show} that the transition
sits near $S = P_T$ and that regret decays as $1/S$; because they run at
fixed finite horizons they cannot distinguish $\Theta(T)$ from $o(T)$, and
because the random-walk comparator is spatially concentrated it could not
certify a budget-induced barrier even asymptotically.
The tight rate above the threshold, and the exact constant at the threshold,
remain \emph{unknown}.

\paragraph{Other limitations.}
The lower bound is one-dimensional; it lifts to $\mathbb{R}^d$ by embedding,
but the dimension dependence of the constants is not resolved.
Our synthetic non-square-wave experiments use quadratic losses, while the
theory covers all convex Lipschitz losses.
We assume full gradient feedback; we note that the square-wave argument is
information-free, so it may transfer to bandit feedback directly, though we
have not verified this.

\section{Conclusion}

A hard movement budget can make dynamic tracking impossible even when the
underlying non-stationarity is mild enough that unconstrained dynamic regret
is sublinear.
Concretely: a deterministic full-range adversary forces linear regret on
every budget-feasible algorithm whenever $S \le cP_T$ for a universal
constant $c>0$ and every $P_T \in [D,DT]$; projected OGD
with a budget-aware step size achieves $o(GDT)$ once $S = \omega(D+P_T)$;
together these identify $D+P_T$ as the critical movement scale, with the
constant-factor band left open.
Above the threshold, budget-controlled OGD gives a concrete $1/S$ upper
bound, but its minimax tightness remains open.
Experiments support the transition picture, locate it near $S = P_T$, and
exhibit empirical $1/S$ decay.

The construction turns on treating comparator range rather than path length
as the relevant measure of difficulty, and we expect that distinction to be
useful wherever a learner is charged for the distance it travels.

\section{Proof of Theorem~\ref{thm:threshold}}
\label{app:thm1}

We prove the two directions separately.
The construction is deterministic, so no expectations are needed.

\subsection{Impossibility}

Fix $D \le P_T \le DT$, take $\Kset = [0,D] \subset \mathbb{R}$ and the losses
\begin{equation}
f_t(x)=G|x-u_t|,
\label{eq:sqloss}
\end{equation}
which are convex, $G$-Lipschitz, and satisfy $f_t(u_t)=0$.
Since these are a special case of the convex $G$-Lipschitz class, a lower
bound against them applies to the whole class.

\paragraph{The comparator.}
Let $n=\min\{\lfloor P_T/D\rfloor,T-1\}\ge 1$ and
$\tau=\lfloor T/(n+1)\rfloor\ge 1$.
For the first $(n+1)\tau$ rounds set
\begin{equation}
u_t =
\begin{cases}
0 & \lfloor (t-1)/\tau\rfloor \text{ even},\\
D & \lfloor (t-1)/\tau\rfloor \text{ odd},
\end{cases}
\end{equation}
and hold the final value for any leftover rounds.
Thus the comparator alternates between the endpoints of $\Kset$ for $n+1$
blocks.
There are exactly $n$ switches, each of length $D$, so the realized path
length is $nD \le P_T$ and the instance is admissible.

\paragraph{Per-block-pair error.}
Group the first $n+1$ blocks into disjoint consecutive pairs
$(B_1,B_2),(B_3,B_4),\dots$,
with $u \equiv 0$ on odd-indexed and $u \equiv D$ on even-indexed blocks.
Let $N = \lfloor (n+1)/2\rfloor$ be the number of complete pairs.
For the $i$-th pair put
$M_i = \sum_{t \in B_{2i-1}\cup B_{2i}}\|x_t - x_{t-1}\|$
and $m_i = \min_{t \in B_{2i-1}\cup B_{2i}} x_t$.
The range of a trajectory is at most its total variation, so every $x_t$ in
the pair lies in $[m_i, m_i+M_i]$.
Hence
\begin{equation}
\sum_{t \in B_{2i-1}}|x_t - 0| \;\ge\; \tau m_i,
\qquad
\sum_{t \in B_{2i}}|x_t - D| \;\ge\; \tau(D - m_i - M_i),
\end{equation}
and adding gives
\begin{equation}
\sum_{t \in B_{2i-1}\cup B_{2i}}|x_t-u_t| \ge \tau(D-M_i).
\label{eq:pair}
\end{equation}
The right-hand side may be negative, in which case the inequality is vacuous
but still valid, which is all we use.

\paragraph{Where the constant is lost.}
Inequality \eqref{eq:pair} bounds the two blocks by a common worst case, and
this is where the factor of two enters.
An exact dynamic program over discretized position and budget, which lower
bounds every algorithm whether online or not, puts the true optimum slightly
below $\tfrac{GDT}{2}(1-S/P_T)$: at $T=240$ and $P_T=29$ it gives $0.364$
versus $0.375$ at $S=P_T/4$, and $0.242$ versus $0.250$ at $S=P_T/2$, in
units of $GDT$.
Crucially it vanishes only at $S=P_T$, not at $S=P_T/2$.
Sharpening this step would recover the constant without changing the
structure of the argument.

\paragraph{Summing.}
The pairs occupy disjoint sets of rounds, so
$\sum_{i=1}^N M_i \le \SC \le S$.
Summing \eqref{eq:pair} over the pairs,
\begin{equation}
\sum_{t=1}^T |x_t - u_t| \;\ge\; \tau\sum_{i=1}^N (D - M_i)
\;\ge\; \tau\,(ND - S).
\end{equation}
Since $N\ge n/2$ and $\tau\ge T/(2(n+1))$, while
$n+1\le 2P_T/D$ whenever $P_T\ge D$, we have
$\tau ND \ge TD/8$.
Also, because $n+1\ge P_T/D$, we have $\tau \le TD/P_T$.
Substituting these two estimates,
\begin{equation}
\Rd = G\sum_{t=1}^T|x_t-u_t|
\;\ge\; \frac{GDT}{8} - \frac{GDT}{P_T}S
\;=\; \frac{GDT}{8}\left(1-\frac{8S}{P_T}\right),
\end{equation}
which is \eqref{eq:lb} with universal constants.
In particular $\Rd = \Omega(GDT)$ whenever $S \le cP_T$ for a sufficiently
small universal constant $c>0$.

\subsection{Achievability}

Run projected OGD with the fixed step size $\eta = S/(GT)$.
Each step moves at most $\eta\|\nabla f_t\| \le \eta G$, so
$\SC \le T\eta G = S$ and the constraint holds by construction.
By Lemma~\ref{lem:zink},
$\Rd \le D(D+2P_T)/(2\eta) + \eta G^2T/2$, which on substituting $\eta$
gives \eqref{eq:ub-constrained}.
If $S = \omega(D+P_T)$ the first term is $o(GDT)$, and if $S = o(DT)$ the
second is $o(GDT)$.
Hence $\Rd = o(GDT)$.

Combining, regret is linear whenever $S \le cP_T$ for a universal constant
$c>0$ and sublinear whenever
$S = \omega(D+P_T)$ with $S = o(DT)$.
Because $P_T\ge D$, both sides identify $P_T$ as the asymptotic movement
scale, while the exact constant-factor band remains unresolved.
\hfill$\square$

\begin{remark}[Why a deterministic adversary]
\label{rem:whydet}
A random-walk comparator $u_t = u_{t-1} + \xi_t$ with $\xi_t$ Rademacher of
scale $\sigma$ has path length $T\sigma$ but spatial range only
$\Theta(\sigma\sqrt T)$, so a learner that never moves and sits at the centre
of the range incurs $\Theta(\sigma\sqrt T)$ per round rather than
$\Theta(D)$.
The resulting bound is $\Omega(P_T^2/T)$, linear only when
$P_T = \Theta(T)$, a regime in which the unconstrained rate is already
linear, so no budget-induced barrier can be concluded from that construction.
The square wave forces the comparator to traverse the entire domain, making
range and diameter coincide.
\end{remark}

\section{Proof of Theorem~\ref{thm:rates}}
\label{app:rates}

We first restate the classical bound we build on.

\begin{lemma}[Dynamic regret of OGD; \citealp{zinkevich2003online}]
\label{lem:zink}
Let $\Kset$ have diameter $D$ and let each $f_t$ be convex and
$G$-Lipschitz.
Projected OGD with fixed step size $\eta>0$ satisfies, for any comparator
sequence with path length $P_T$,
\begin{equation}
\sum_{t=1}^T\big[f_t(x_t)-f_t(u_t)\big]
\;\le\; \frac{D(D+2P_T)}{2\eta} + \frac{\eta G^2 T}{2}.
\end{equation}
\end{lemma}

\paragraph{Upper bound.}
Run projected OGD with the fixed, time-independent step size
$\eta = \min(\sqrt{D(D+2P_T)/(G^2T)},\, S/(TG))$.
Per-round movement is at most $\eta G$, so $\SC \le T\eta G \le S$.

\emph{Case 1 (budget-rich).}
When $S \ge \sqrt{DT(D+2P_T)}$ we have $\eta = \eta^\star =
\sqrt{D(D+2P_T)/(G^2T)}$, and Lemma~\ref{lem:zink} gives
$\Rd \le G\sqrt{D(D+2P_T)T}$.
The dependence on the diameter here is $\sqrt D$ rather than $D$: minimizing
$D(D+2P_T)/(2\eta) + \eta G^2T/2$ yields
$G\sqrt{D(D+2P_T)T} \approx G\sqrt{2D}\sqrt{P_TT}$ when $P_T \ge D$.
We keep the exact form throughout.

\emph{Case 2 (budget-constrained).}
When $S < \sqrt{DT(D+2P_T)}$ we have $\eta = S/(TG)$, and substituting,
\begin{equation}
\frac{D(D+2P_T)}{2\eta} = \frac{GTD(D+2P_T)}{2S},
\qquad
\frac{\eta G^2T}{2} = \frac{GS}{2},
\end{equation}
so $\Rd \le O(GTD(D+2P_T)/S)$.
Taking the minimum with the trivial bound $\Rd \le GDT$ gives
\eqref{eq:rate-ub}.
\hfill$\square$

\paragraph{For comparison: unconstrained lower bound.}
$\Rd \ge \Omega(G\sqrt{DP_TT})$ is the unconstrained dynamic regret lower
bound of \citet{besbes2015non}.
Their construction does not rely on a movement constraint, so it remains a
valid lower bound in the budgeted setting whenever the budget is large enough
to permit the comparator-scale motion. We do not provide a matching lower
bound for the $GTD(D+2P_T)/S$ term in \eqref{eq:rate-ub}; this is the main
open rate question left by the paper.

\section{Extensions}
\label{app:extensions}

The following are corollaries of Theorem~\ref{thm:threshold}, obtained by
computing an induced path length and invoking the main theorem; they contain
no new argument.
They are stated under the hypothesis $P_T = \Theta(T)$, which is the regime
in which the unconstrained rate is already linear, so on their own they do not
certify a budget-induced barrier.
Applying the square-wave construction in these settings, which would require
exhibiting latent dynamics whose induced comparator traverses the domain, is
left to future work.
We record them as illustrations of the reduction, not as independent
contributions.

\paragraph{Latent dynamical systems.}
Let $\{z_t\}$ evolve as $z_{t+1} = \mathcal{T}(z_t,\xi_t)$ for a possibly
unknown transition operator $\mathcal{T}$ and exogenous noise $\{\xi_t\}$,
with losses $f_t(x) = \ell(x, z_t)$ where $\ell(\cdot,z)$ is convex and
$L$-Lipschitz, and let $u_t \in \arg\min_{x\in\Kset} f_t(x)$.
If the induced path length satisfies $P_T = \Theta(T)$, then any algorithm
with $S = o(T)$ has $\mathbb{E}[\Rd] = \Theta(T)$, while
$S = \Omega(P_T)$ makes sublinear dynamic regret achievable.

\paragraph{Event-driven environments.}
When environment changes arrive as a point process~\citep{daley2007introduction},
the induced path length is governed by the expected number of events, and the
same reduction applies once that quantity is $\Theta(T)$.

\paragraph{Spectral view.}
For environments generated by a linear evolution operator, the induced path
length can be expressed through the operator's
spectrum~\citep{mezic2005spectral,budisic2012applied}, and the feasibility
condition again reduces to comparing $S$ with $P_T$.

\section{Additional Experimental Results}
\label{app:extra-exp}

\paragraph{Setup.}
Synthetic experiments use two comparators, for two different purposes.
The square wave of Appendix~\ref{app:thm1} is the instance the lower bound is
proved against, and is used in Sections~\ref{sec:exp-sqwave} and
\ref{sec:exp-sublinear}.
The random walk $u_t = u_{t-1} + \xi_t$ with $\xi_t$ Rademacher is used for
the transition-location and $1/S$ plots; as Remark~\ref{rem:whydet} explains
it cannot certify a budget-induced barrier and we do not use it for that.
Unless stated otherwise, results are averaged over 20 seeds.

\begin{figure}[h]
\centering
\includegraphics[width=0.85\textwidth]{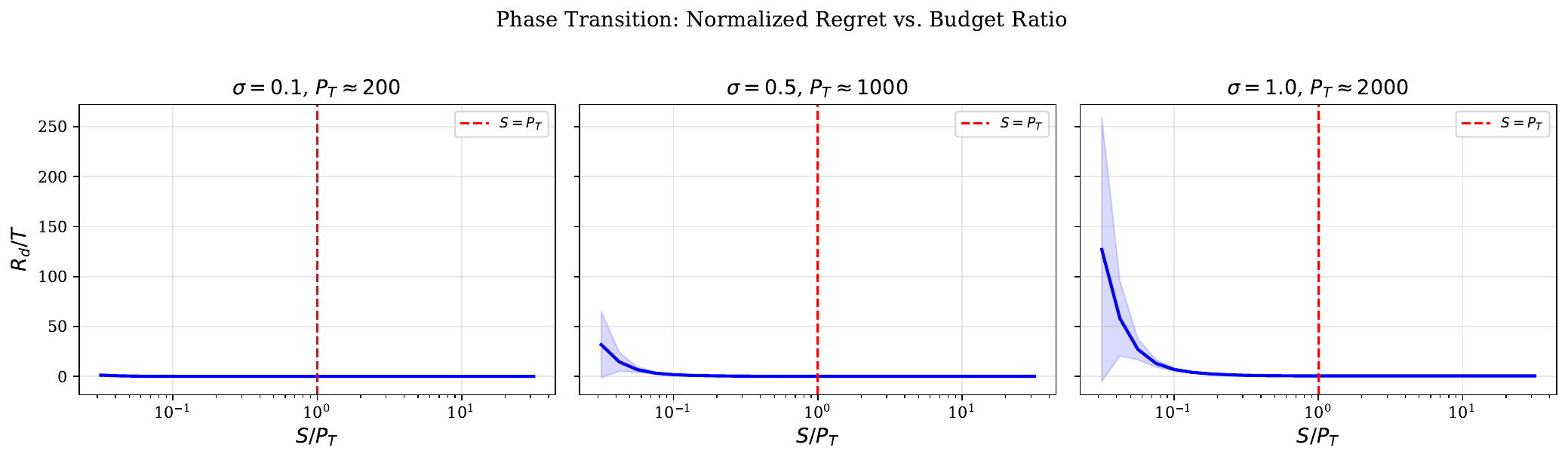}
\caption{Random-walk comparator: normalized regret $R_d/T$ against $S/P_T$
for drift magnitudes $\sigma \in \{0.1,0.5,1.0\}$ at $T=5000$, with the red
dashed line at $S=P_T$ and shading showing $\pm1$ standard deviation over
20 seeds.
The drop near $S=P_T$ locates the transition; because this comparator is
spatially concentrated, it does not by itself certify a budget-induced
barrier (Remark~\ref{rem:whydet}).}
\label{fig:phase-rw}
\end{figure}

\begin{figure}[h]
\centering
\begin{minipage}[t]{0.48\textwidth}
\centering
\includegraphics[width=\textwidth]{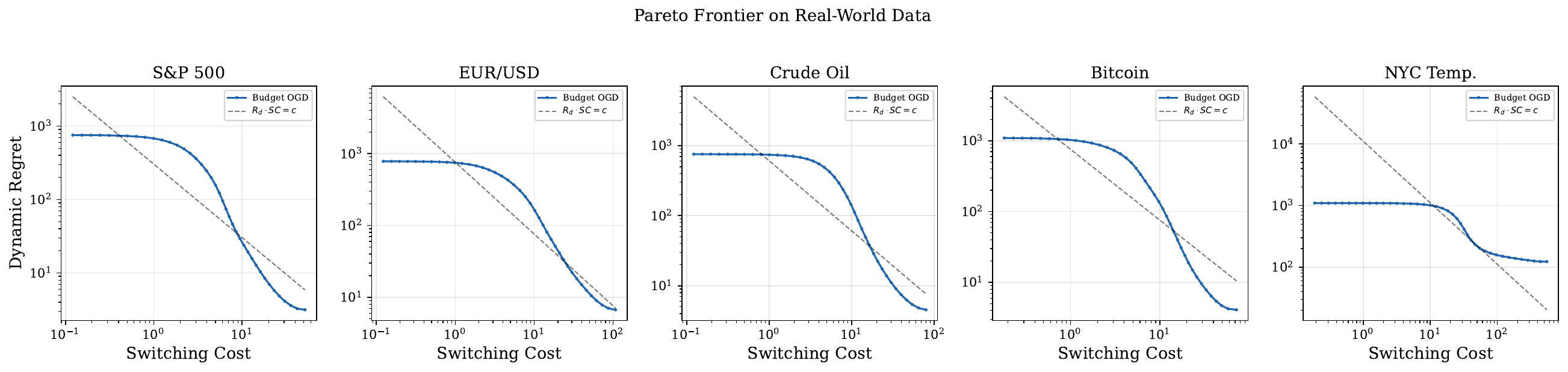}
\caption{Budget--regret trade-off on the five real series.
The curves follow $R_d \propto 1/\SC$ over the mid-range of budgets,
consistent with the upper bound \eqref{eq:rate-ub} only.}
\label{fig:realpareto}
\end{minipage}\hfill
\begin{minipage}[t]{0.48\textwidth}
\centering
\includegraphics[width=\textwidth]{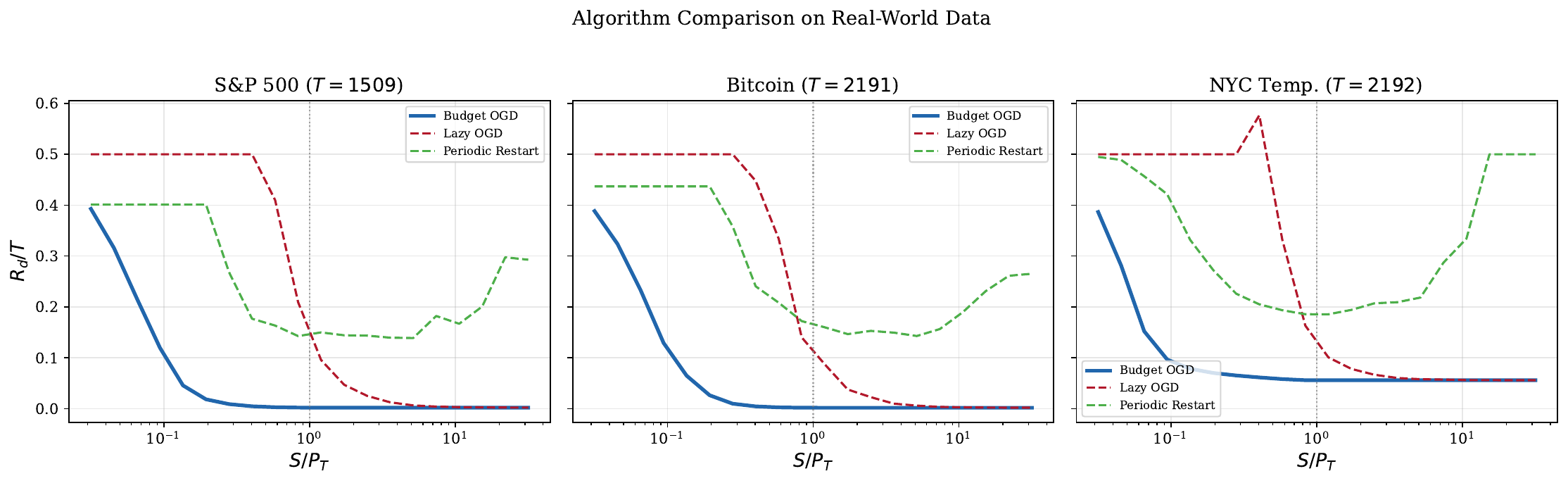}
\caption{Algorithm comparison on real series.
Budget-Controlled OGD dominates Lazy OGD and Periodic Restart on the
S\&P~500, Bitcoin, and NYC temperature.}
\label{fig:realalgo}
\end{minipage}
\end{figure}

\begin{figure}[h]
\centering
\includegraphics[width=0.85\textwidth]{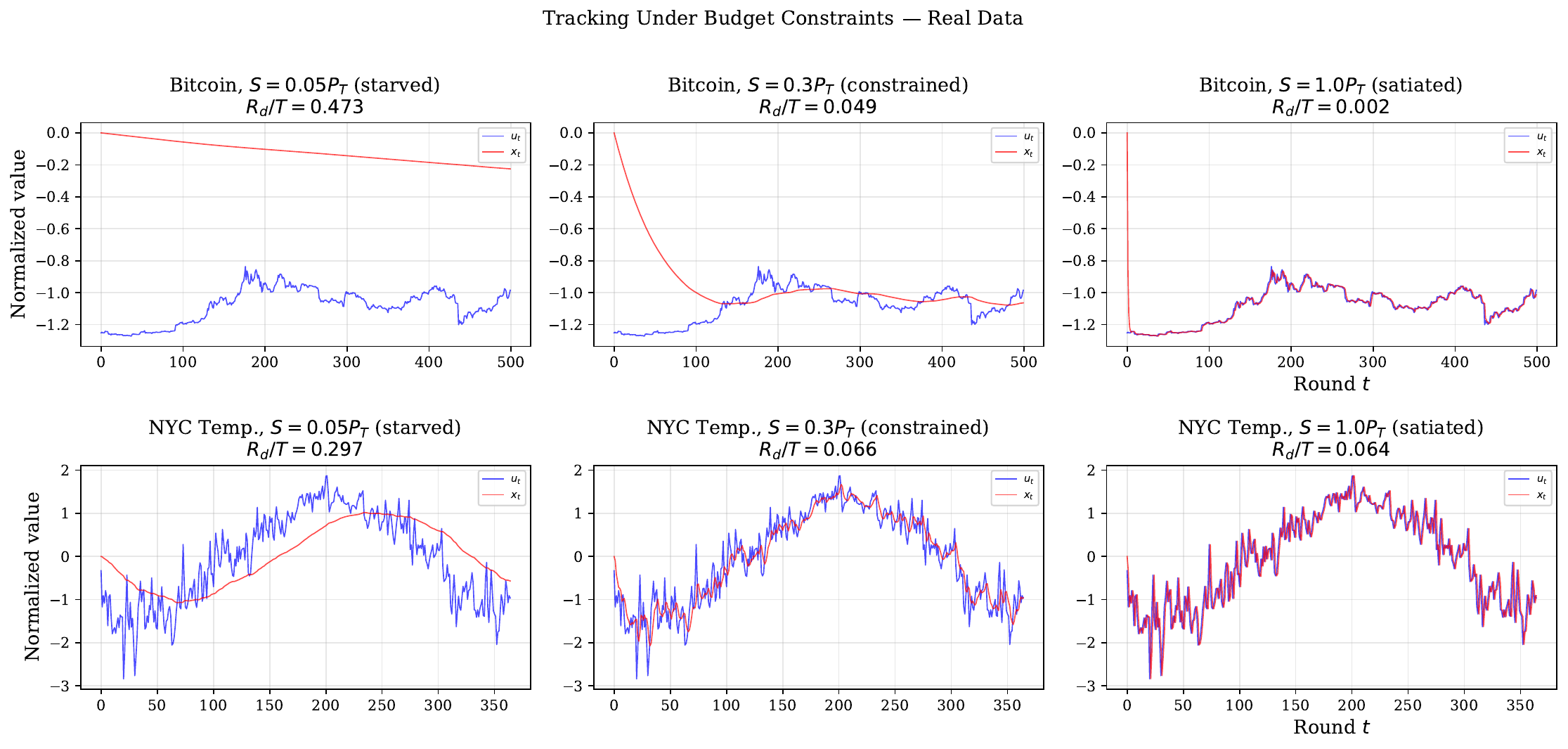}
\caption{Learner trajectories on Bitcoin and NYC temperature at three budget
levels.
At $S=0.05P_T$ the learner barely moves and incurs $R_d/T \approx 0.47$;
at $S=0.3P_T$ tracking improves substantially; at $S=P_T$ the learner tracks
near-perfectly with $R_d/T < 0.01$.}
\label{fig:realtraj}
\end{figure}

\end{document}